\documentclass[conference]{IEEEtran}
\usepackage{times}

\usepackage[numbers]{natbib}
\usepackage{multicol}
\usepackage[bookmarks=true]{hyperref}
\usepackage{cite}
\usepackage{amsmath,amssymb,amsfonts}
\usepackage{caption}
\usepackage{ulem}
\usepackage{subcaption}
\usepackage{graphicx}
\usepackage{textcomp}
\usepackage{xcolor}
\usepackage{float}
\usepackage{todonotes}
\usepackage{xcolor}
\usepackage{bm}
\usepackage{amsmath}
\usepackage{algorithm}
\usepackage{algpseudocode}
\usepackage{stfloats}
\renewcommand{\vec}[1]{\bm{#1}}
\newcommand{\mat}[1]{\bold{#1}}

\begin{document}

\title{Vib2Move: In-Hand Object Reconfiguration via Fingertip Micro-Vibrations}

% You will get a Paper-ID when submitting a pdf file to the conference system
% \author{Author Names Omitted for Anonymous Review. Paper-ID 346}

\author{\authorblockN{Xili Yi}
\authorblockA{Department of Robotics\\
University of Michigan\\
Michigan, Ann Arbor 48109\\
Email: yixili@umich.edu}
\and
\authorblockN{Nima Fazeli}
\authorblockA{Department of Robotics\\
University of Michigan\\
Michigan, Ann Arbor 48109\\
Email: nfz@umich.edu}}

\maketitle

\begin{abstract}

We introduce Vib2Move, a novel approach for in-hand object reconfiguration that uses fingertip micro-vibrations and gravity to precisely reposition planar objects. Our framework comprises three key innovations. First, we design a vibration-based actuator that dynamically modulates the effective finger–object friction coefficient, effectively emulating changes in gripping force. Second, we derive a sliding motion model for objects clamped in a parallel gripper with two symmetric, variable-friction contact patches. Third, we propose a motion planner that coordinates end-effector finger trajectories and fingertip vibrations to achieve the desired object pose. In real-world trials, Vib2Move consistently yields final positioning errors below 6 mm, demonstrating reliable, high-precision manipulation across a variety of planar objects.
For more results and information, please visit  \href{https://vib2move.github.io}{https://vib2move.github.io}.

% In this paper, we address the mechanics and planning algorithms required to slide an object to an arbitrary pose using a combination of micro-vibrations from the fingertips and gravity. We introduce a novel vibration-based gripper actuator capable of dynamically modulating the equivalent friction coefficient with a high response rate. Additionally, we discuss the underlying mechanism and its potential to emulate equivalent gripping force control. Leveraging this capability, we derive a sliding motion model for planar objects held by a parallel gripper with two symmetric contact patches. Based on this analysis, we propose a heuristic method for in-hand planar object manipulation and validate it through real-world experiments. Our results demonstrate that the proposed approach enables accurate object positioning across various planar objects, achieving final pose errors of less than {TODO} mm.
\end{abstract}

\IEEEpeerreviewmaketitle

\section{Introduction}

In-hand manipulation—the ability to reposition an object within the robot’s grasp—is central to achieving dexterous and versatile robotic applications. Among the myriad of available end-effectors, parallel jaw grippers stand out for their simplicity, low cost, and widespread adoption across various industries. However, their single degree of freedom (DoF) poses a fundamental challenge: it severely restricts the types of in-hand reconfiguration that can be performed. This gap in capability not only limits the range of tasks that can be executed but also underscores the need for new methods to enhance the dexterity of these otherwise robust and practical grippers.

Prior research on in-hand reconfiguration can be broadly categorized into two classes: methods that adapt single DoF end-effectors (primarily parallel jaw grippers) and those that exploit dexterous multi-fingered hands. In the former group, one common strategy is to leverage environmental features, such as walls or shelves, effectively treating them as additional “virtual fingers.” These approaches often rely on known geometry, friction properties, and precise force control to achieve the desired manipulations. While effective under controlled conditions, they falter when the environment is not well-defined or when real-time adjustments in grasping force are required—a nontrivial feat for conventional parallel jaw grippers. Meanwhile, multi-fingered robotic hands circumvent many of these issues by offering richer manipulation capabilities, but their inherent complexity, cost, and calibration overhead often make them impractical for industrial contexts. As a result, a method that marries the simplicity of a two-finger design with the dexterity of more advanced end-effectors remains highly desirable.

\begin{figure}[t]
    \centering
    \includegraphics[width=0.9\linewidth]{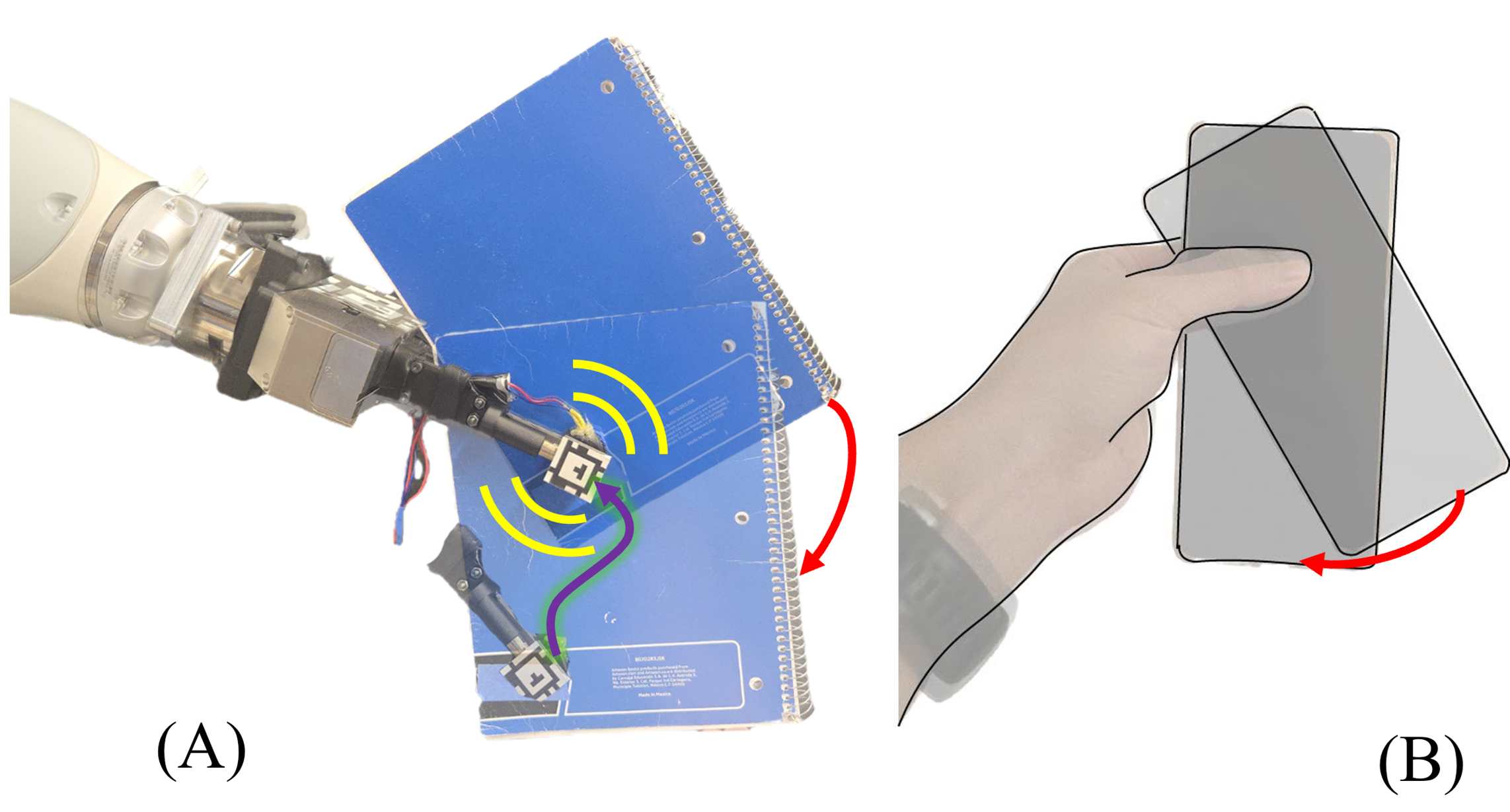}
    \caption{(a) Robot pivots a notebook in free space, using gravity and fingertip vibrations. (b) Human pivoting a cellphone in free space, utilizing gravity and precise control of grasping force}
    % \textcolor{red}{I would zoom into the robot hand a bit more, and can we add like an additional image illustrating the fact that the fingers vibrate? Maybe a picture of the end-effector with some sort of arrows or squiggly lines to illustrate what vibrates?}.}
    \label{fig:teaser}
\end{figure}

To address this need, we propose a vibration-based gripper design that dynamically modulates the effective friction between the fingertips and the grasped object. By introducing micro-vibrations at the contact interface, our design induces rapid transitions between stick and slip regimes, effectively providing high-frequency control over the “equivalent” grasping force. Building on this mechanism, we develop a friction and motion model specifically for planar objects grasped vertically in free space, where gravity is the sole external force. Unlike previous approaches requiring detailed knowledge of contact properties or object mass, our model depends only on the geometry of the finger contact area. This characteristic grants a notable degree of robustness by tolerating uncertainties in friction and object weight, thus avoiding the fine-tuned parameter calibration that often hampers manipulation tasks. While we demonstrate our approach on a parallel jaw gripper, the underlying principle of vibration-based friction modulation can seamlessly extend to more sophisticated or multi-fingered end-effectors, broadening its applicability to a wider range of hardware platforms.

We implement this approach on a parallel jaw gripper outfitted with low-profile vibration motors in each finger, employing only on-off vibration signals for rapid friction control. By integrating a heuristic-based subgoal generation scheme and a closed-loop visual controller, our system can reposition planar objects in free space without specialized fixtures or intricate force sensors. Prior work has explored environmental-assisted manipulation \citep{chavan2020planar} and multi-DoF vibration-based end-effectors \citep{nahum2022robotic, binyamin2024vibration}, but these approaches demand additional resources, known geometry, or more complex hardware. In contrast, our design retains the inherent simplicity and reliability of standard parallel jaw grippers while offering a versatile method for in-hand reconfiguration.

\subsection{Problem Statement}

\begin{figure}
    \centering
    \includegraphics[width=0.7\linewidth]{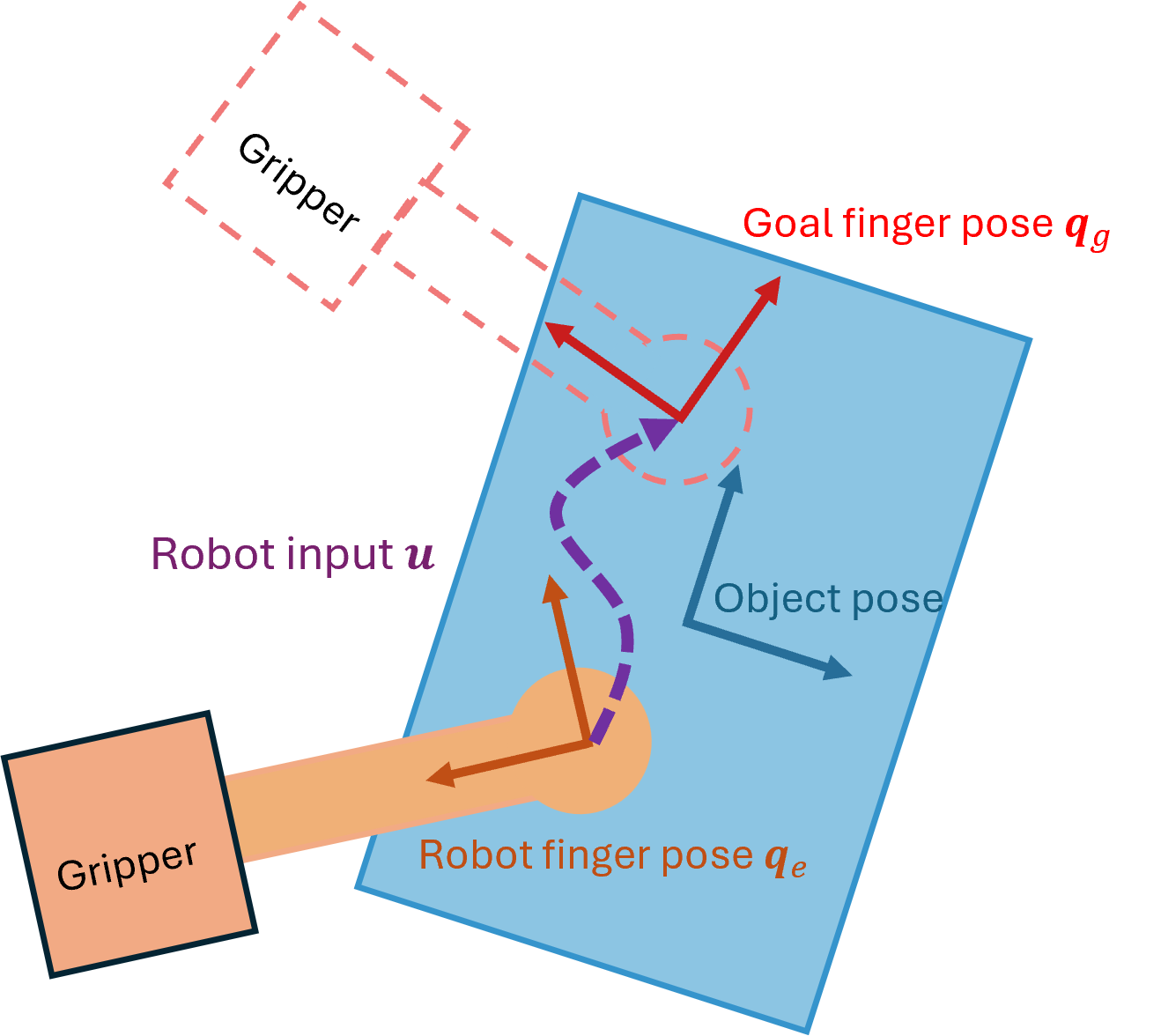}
    \caption{Problem statement: Given the robot finger pose $\vec{q}_{e}$, find the robot input $\vec{u}$ to drive the finger to goal pose $\vec{q}_{g}$. }
    % \textcolor{red}{The figure is super blurry, is there a higher resolution version? The gripper visualizations are not very pretty, if we have time, I'd suggest either using a figure of the real robot with the same annotations or making it a little prettier.}}
    \label{fig:problem_statement}
\end{figure}

Consider the exemplar task illustrated in Fig. \ref{fig:problem_statement}. The goal of the robot is to move the object from its initial configuration in the grasp to the desired configuration. The object motion is constrained to the vertical plane, where gravity acts exclusively in the negative direction. We assume that the grasping contact remains a patch contact with a constant shape and that the friction is homogeneous. We also assume both the object and the gripper are rigid.

% \todo{rewrite this}
Let $\vec{q}_e^w(t) \in \mathrm{SE}(2)$ 
and $\vec{q}_o^w(t) \in \mathrm{SE}(2)$  represent the poses of the robot finger and the object in world frame, at time $t$ respectively. We define \(\vec{q}_{r}^o\) $\in$ \(\mathrm{SE}(2)\) as the relative pose of the robot finger with respect to the object frame, given by $ \vec{q}_{r}^o(t)=\mat{J_B}( \vec{q}_e^o(t) - \vec{q}_o^o(t)) $, as illustrated in Fig. \ref{fig:problem_statement}.
The relative pose $ \vec{q}_{r}^o(t) $, deviates from its initial value only due to slip between the finger and the object. Given an arbitrary relative robot finger goal pose \(\vec{q}_g^o \in \mathrm{SE}(2)\) for the robot finger to reach on the object, our objective is to design an input consisting of the end-effector finger trajectory and vibration control input \(\vec{u}(t)=[\vec{q}_e^w(t), v_e(t)]\), where $v_e(t)$ is vibration control, % over time step \(t = {1,2,..., n}\) 
to minimize the final pose error:
\begin{equation*}
\vec{u}^* = \arg \min_{\vec{u}} ||\vec{q}_{e}^o(T) - \vec{q}_g^o ||\label{eq:ctrlprob}
\end{equation*}
where \(T\) is time at the end of the path. 
To achieve the goal pose, it is necessary to regulate the relative slippage, as the relative pose is controlled through this slippage.

\section{Related works}

In-hand manipulation has been extensively studied, encompassing both prehensile and non-prehensile strategies. Many works focus on leveraging high-DoF robotic hands for dexterous in-hand manipulation. Both model-based and learning-based approaches have significantly advanced these strategies. For instance, object-level impedance control ensures compliant 6-DoF positioning while maintaining grasp stability during contact reconfigurations, such as finger gaiting \citep{pfanne2020object}. Contact-implicit model predictive control (MPC) enables dynamic replanning under uncertainty without predefined contact sequences \citep{jiang2024contact}. Reinforcement learning (RL) has demonstrated success in handling contact variability and generalizing across objects \citep{luo2024progressive, chen2022system}, while hybrid frameworks combine RL for planning with model-based controllers for stability \citep{zarrin2023hybrid, veiga2020hierarchical, arunachalam2023dexterous}. Additionally, motion primitive dictionaries derived from human demonstrations simplify modeling by indirectly capturing manipulation constraints \citep{solak2019learning, hammoud2024robotic}.
Parallel grippers have also been studied for in-hand manipulation, primarily with planar objects. Due to their single degree of freedom, their manipulation capabilities are inherently limited. Karayiannidis et al. \citep{karayiannidis2016adaptive} proposed an adaptive control approach for pivoting with a parallel gripper using both visual and tactile feedback. However, their method focused solely on orientation and was constrained to cases with gravity-induced torque. Chavan-Dafle et al. \citep{chavan2020planar} introduced motion cones, leveraging the environment as an external pusher for manipulating planar objects. In our work, we do not depend on external features. Hou et al. \citep{hou2018fast} designed a unique locking mechanism attached to the fingertips and two primitives to either roll or pivot an object in the grasp. While offering significant dexterity, this method also requires using the environment.

To address the limitations of parallel grippers, researchers have explored modifications to introduce additional DoFs. For instance, Ma et al. \citep{ma2016hand} incorporated a conveyor-like surface into the gripper to enhance its in-hand manipulation capabilities. Nahum et al. \citep{nahum2022robotic} and subsequent work by Binyamin et al. \citep{binyamin2024vibration} added rotation and vibration to the gripper fingertips, enabling arbitrary planar object manipulation.
In our work, we employ a commercial parallel gripper and augment it with vibrational fingertips, achieving arbitrary pose manipulation of planar objects.

The modeling, planning, and control of planar object pushing or sliding have been extensively studied. Stuber et al. \citep{stuber2020let} conducted a comprehensive survey on robotic pushing, covering both analytical and data-driven approaches. Lynch and Mason \citep{lynch1996stable} proposed a motion planning algorithm leveraging sticking interactions to achieve target trajectories. These manipulation strategies—often referred to as dragging or sliding—rely on quasi-static analysis, Coulomb friction with limit surface modeling, and soft contacts for moment transfer.
Kao et al. \citep{kao1992quasistatic} examined compliance and sliding mechanics, which were later extended to the manipulation of lightweight objects, such as a business card, using two sliding robot fingers on a frictionless table \citep{kao1993comparison, xue1994dexterous}. However, such frictionless assumptions are impractical for heavier objects. Ghazaei et al. \citep{ghazaei2020quasi} introduced a hybrid dynamical system to predict and control object motion under conditions such as sticking, slipping, and pivoting. While this work addressed some dynamic challenges, the associated planning problem remains unresolved.
Efforts to minimize unintended slippage \citep{fakhari2019modeling} and to develop model-free dragging methods for objects with unknown dynamics \citep{9811686} have demonstrated potential but are generally constrained to planar translation or simulation environments. More recent studies by Yi et al. \citep{yi2023precise, yi2024dual} advanced planar sliding using horizontal or tilted supporting surfaces, further pushing the boundaries of the field.
While these works mark significant progress, they also underscore the challenges of extending such methods to more complex and dynamic tasks.

% survey about data-driven / model-based, data-driven based to learn shape, unknown mass..., Goyal e contact between a robot finger and an object during slippage under both torsion and shearfriction model for planar contact -LS, simplified LS, sliding dynamics, planar sliding with top contact from Yi, for both horizontal and tilted. 

Vibration has long been considered an effective form of actuation and has been widely deployed in applications such as conveyor belts. In the field of topography, the friction-reducing effects of vibration are well-documented. Numerous analytical \citep{liu2020analytical} and experimental \citep{holl2018experimental} studies have provided explanations for this phenomenon. Due to its simple structure and ease of activation, vibration has found extensive use in industry, particularly in conveyor belts \citep{winkler1978analysing}.
In recent years, roboticists have also explored the use of vibration in robotics. The stick-slip effect \citep{gao1994dynamic} forms the basis for vibration-based manipulation, where the contact interface alternates between adhering through static friction (with no movement) and sliding due to kinetic friction. In \citep{rubenstein2014kilobot, rubenstein2014programmable}, the authors leveraged this stick-slip effect to develop the Kilobot, a swarm robotics platform that achieves effective locomotion.
Several works have focused on integrating vibration functionality into robot grippers. A recent study and its follow-up \citep{nahum2022robotic, binyamin2024vibration} introduced vibration to in-hand manipulation. They proposed the Vibratory Finger Manipulator (VFM), a novel mechanism featuring a rotary finger actuator embedded with a vibration motor to manipulate planar objects. However, this approach is limited to horizontal planes, and the gripper design is complex, requiring additional degrees of freedom (DoFs). \citet{maruo2022dynamic} presented a dynamic underactuated manipulator that exploits structural anisotropy in a spiral flexible body to achieve multi-DoF object motion through vibration-induced orbit modulation using a single actuator. More recently, \citet{yako2024vertical} demonstrated vertical vibratory transport of grasped parts by leveraging impact-induced accelerations combined with periodic stick-slip motion, enabling transport against gravity.

In this work, we propose a simple vibration-based parallel gripper finger design that enables precise and rapid control of the equivalent grasping force to regulate slippage. We also derive the mechanics and motion model for grasping a planar object in a vertical plane, where gravity is the only external force and the contact is modeled as a patch contact. Based on this, we propose a planning and control algorithm to manipulate the object in-hand.

\section{Mechanics}
To achieve precise in-hand object reconfiguration, it is essential to understand the motions induced by the vibration-based gripper in this underactuated system. This understanding forms the foundation for to our planning and control algorithm approach. Given the object pose $\vec{q}_o^w(t)=[x_o^w(t), y_o^w(t), \theta_o^w(t)]^T$ and the robot finger pose $\vec{q}_e^w(t)=[x_e^w(t), y_e^w(t), \theta_e^w(t)]^T$, we assume that when the fingers begin vibrating, the robot finger remains fixed. The objective is to predict the motion of the object $\{\vec{q}_o^{w}(t), \vec{q}_o^{w}(t+1), \dots, \vec{q}_o^{w}(t+n)\}$ , which is governed by the interaction between friction, vibration, and gravity. This section introduces the limit surface friction model, discusses possible motions during reconfiguration, and examines the impact of vibration on frictional interactions.

\subsection{Free-space Sliding with Patch Contact}
\subsubsection{Limit Surface}
The limit surface defines the boundary of all possible frictional wrenches a contact patch can exert on the object. It also characterizes the set of instantaneous object twists due to the frictional interaction at the contact \citep{howe1988sliding,goyal1989planar}.   

\begin{figure}
    \centering
    \includegraphics[width=0.9\linewidth]{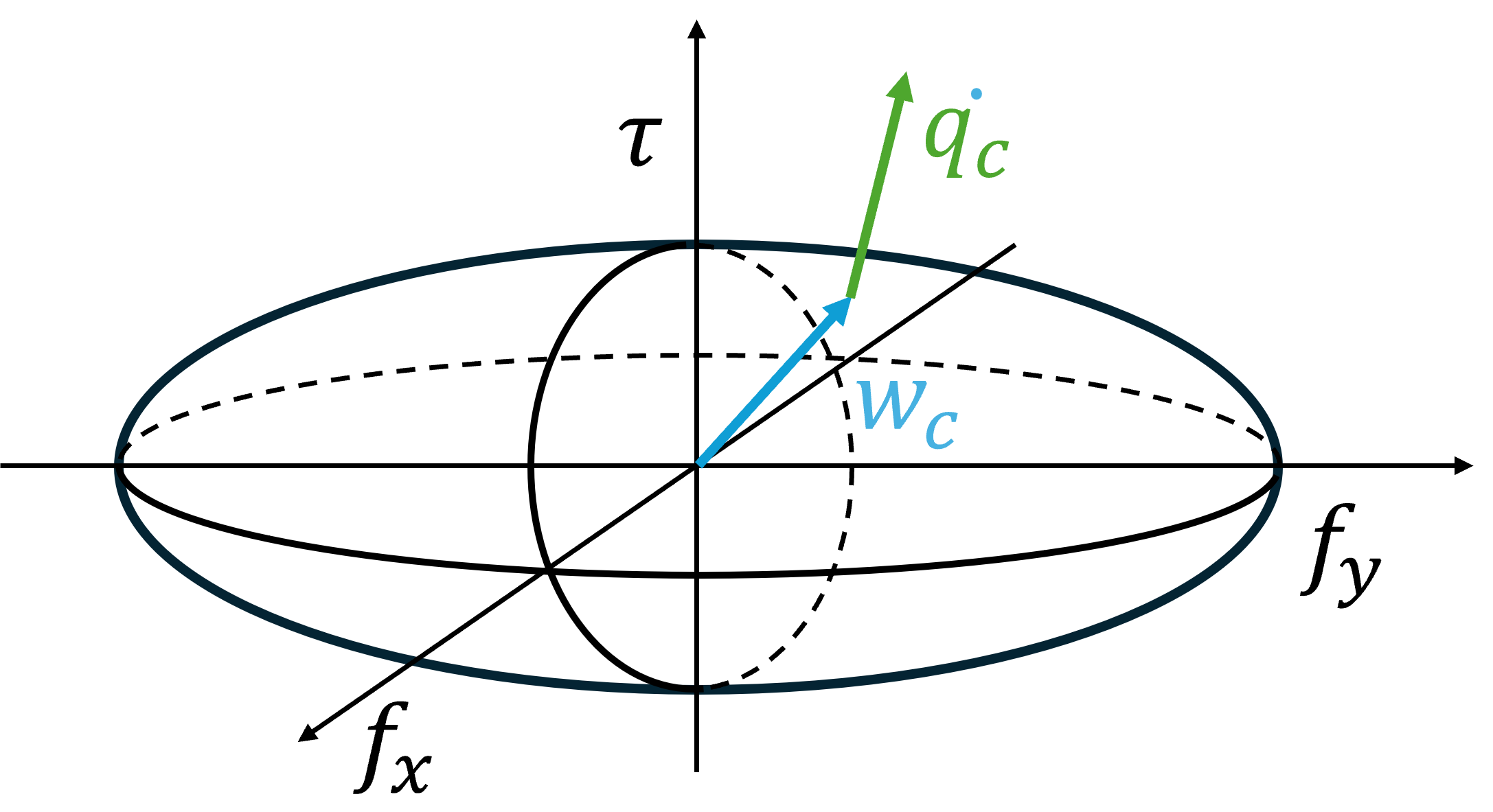}
    \caption{Limit surface is defined as the boundary of the set of all possible frictional wrenches that a supporting contact patch can offer.}
    \label{fig:limit_surface}
\end{figure}

In general, calculating the limit surface in closed form is impossible; however, the seminal work of \citet{howe1996practical} proposed an ellipsoidal approximation in the wrench space that has proved effective in a variety of subsequent studies \citep{lynch1996stable,xydas1999modeling,hogan2020feedback,hogan2020reactive,kloss2022combining,shi2017dynamic} and forms the basis of our analysis.

Let $\vec{w}=[f_x, f_y, \tau_z]^T$ denote the friction wrench on the object from the supporting contact in the contact frame. The ellipsoidal limit surface can be expressed as:
\begin{equation*}
    % \frac{f_x^2}{a_1}+\frac{f_y^2}{a_2}+\frac{m_z^2}{a_3} =1
    \vec{w}^T\mat{A}\vec{w}=1
    \label{eqn:limit_surface}
\end{equation*}
where $\mat{A}=\text{Diag}\{(a_1N)^{-2}, (a_2N)^{-2}, (a_3N)^{-2}\}$, $N$ is the normal force, and $a_1$, $a_2$, $a_3$ are geometric and frictional coefficients.
Assuming isotropic Coulomb friction model, we have $a_1=a_2=\mu$, where $\mu$ is the friction coefficient between the contact and the object. We can represent arbitrary contact patches using their equivalent radius \(r_o\) \citep{howe1996practical}. Using this radius, we can write the maximum friction torque about the contact normal as $a_3=r_0c\mu$, where $r_0$ is the equivalent radius of the object and $c\in [0,1]$ is a constant corresponding to object geometry. This constant is obtained by integration and for a uniform pressure distribution with the equivalent radius contact, $c \approx 0.6$ \citep{xydas1999modeling, shi2017dynamic}. 
We can now write the relationship $ a_3 = rca_2=rca_1 $ where we note that for a fixed contact patch size, $A\propto N$, -- i.e., that the size of the limit surface is proportional to the normal force while maintaining the limit surface axes ratios.

As is standard with the ellipsoidal limit surface approximation \citep{fakhari2019modeling,hogan2020feedback,zhou2016convex, 9811686,ghazaei2020quasi}, we assume that the maximum static friction is equal to the kinetic friction under the same normal force. Static friction wrenches are inside or on the limit surface, while dynamic friction wrenches are strictly on the limit surface. When the object slides on the contact plane, the wrench $\vec{w}_c$ lies on the limit surface and the surface normal at this point is the direction of instantaneous object twist $\vec{q}_c$ \citep{howe1996practical}:% at the contact based on the maximal energy dissipation principle. 
\begin{equation}
    \dot{\vec{q}_c} \parallel \mat{A} \vec{w}_c
    \label{eqn:ls_twsit_direction}
\end{equation}

\subsubsection{Sliding motion model}
\begin{figure}
    \centering
    \includegraphics[width=0.9\linewidth]{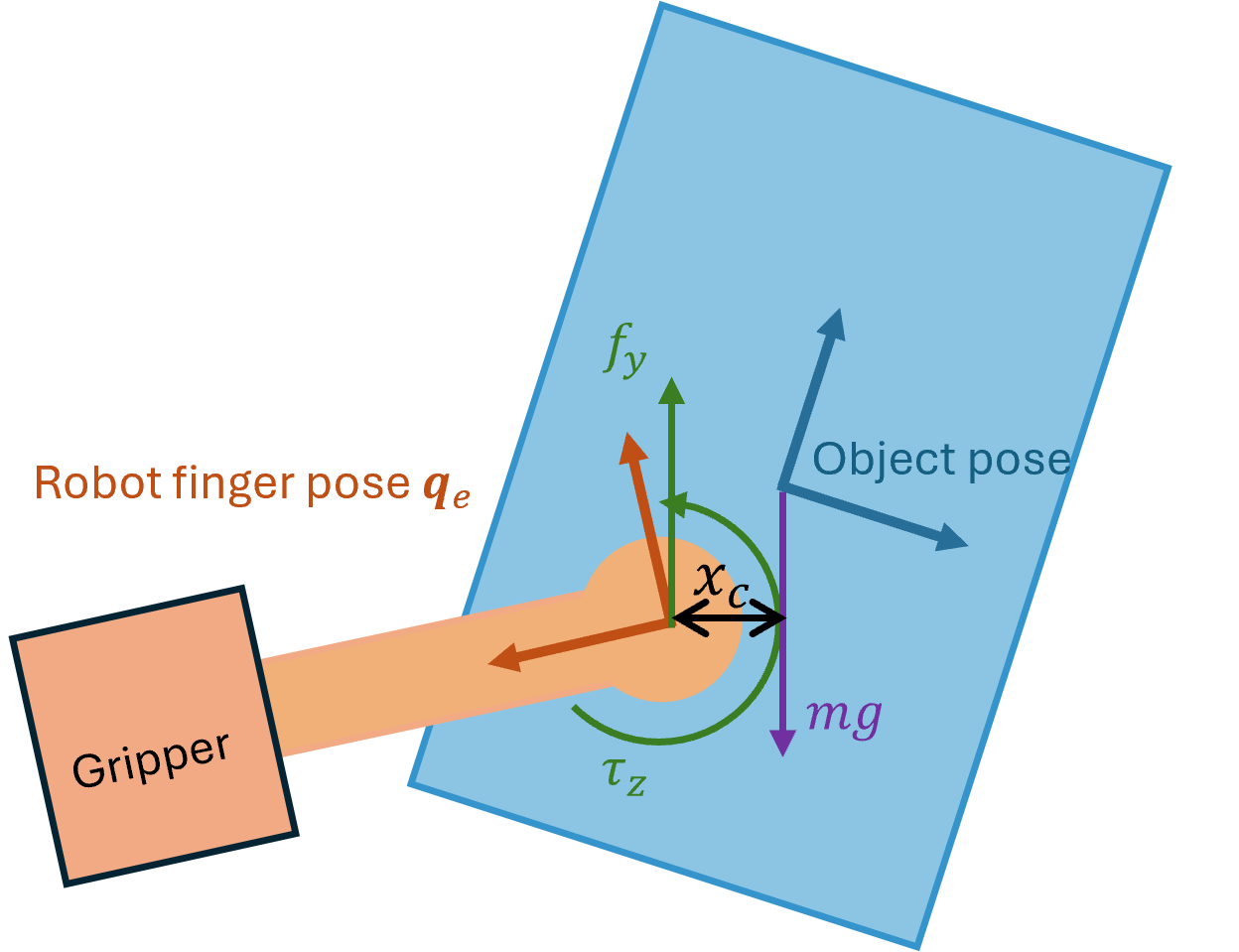}
    \caption{Force-torque balance of the object in the vertical plane. The brown frame indicates the robot finger's contact patch on the object, while the blue frame denotes the object itself. In this configuration, the gravitational force is balanced by the frictional wrench.}
    % \textcolor{red}{I think this figure is not complete yet, I suggest changing the color pallet, the green and red are too harsh. I also think the caption is incomplete}}
    \label{fig:ft_balance}
\end{figure}
In free space in-hand reconfiguration, we model the system in $\mathrm{SE}(2)$, as illustrated in Fig. \ref{fig:ft_balance}. The parallel gripper applies both normal forces and frictional wrenches to the object through the contact patches. When considering force-torque balance in $\mathrm{SE}(2)$, we focus solely on the frictional wrenches because the normal forces from the two contact patches cancel each other out. Gravity acts as a point force applied to the object’s center of mass. 

The object's slip with respect to the fingers is induced by vibrations. The object's motion is governed by the net force applied to it through the contact patch and gravity. In our method, we modulate the finger vibrations such that the induced object slip is quasi-static. Here, modulation is achieved by controlling the length of time for which the vibration is applied. The longer the application, the more time the object has to accumulate acceleration. By focusing on short pulses, we induce quasi-static motion which simply means that inertial forces are negligible and the the object comes to rest as soon as vibrations are turned off. Under this assumption, the object always maintains force-torque equilibrium, whether it slides or remains stationary, regardless of the exact friction wrench. We evaluate this assumptions in the results section.

Consequently, given the object’s orientation, grasping contact location, and equivalent diameter, we may write force-torque balance as:
\begin{equation}
    \begin{aligned}
    f_x^w=0\\
    mg + f_y^w=0  \\    
    \tau_f^w+mgx_c^w=0
    \end{aligned}
    \label{eqn:ft_balance}
\end{equation}
where $[f_x^w, f_y^w, \tau_f^w]^T$ is the friction wrench due to the fingers in the world frame, $x_c$ is the distance from object center to finger in x direction. We can rearrange to arrive at:
\begin{equation}
    \begin{aligned}
    f_x^w &=0\\
    f_y^w &=-mg  \\  
    \tau_f^w &=-mgx_c^w
    \end{aligned}
\end{equation}

We observe that the robot finger pose does not affect the object’s motion directly; only the relative pose between the robot finger and the object’s center is relevant. Thus, when planning the robot’s motion, we need only consider the finger’s orientation with respect to the world frame.

Notably, we do not require knowledge of the grasping force or friction coefficient to determine the frictional wrench. Instead, we enforce the balance of the object to obtain this wrench. This choice reflects the fact that the shape of the finger’s limit surface depends solely on the contact’s radius, whereas the normal force and friction coefficient merely scale the size of the limit surface. By enforcing the wrench balance, effectively select the corresponding limit surface size. 

As discussed in the previous subsection, when the wrench lies on limit surface, the object either sticks to the robot finger or slides along the twist direction in Eq. \ref{eqn:ls_twsit_direction}. 
We also assume that whenever vibration is applied, the object must slide; when vibration ceases, the object maintains sticking contact with the robot finger.

Using the limit surface friction model described, we can determine the contact’s motion direction:
\begin{equation}
    \dot{\vec{q}}\parallel \mat{A} \vec{w}_f
    \label{eqn:motion_direction}
\end{equation}
where $\vec{w}_f = [f_x, f_y, \tau_f]^T $ and $\dot{\vec{q}}$ denotes the instantaneous motion of the object. Notably, Eq.~\ref{eqn:motion_direction} indicates only the direction of motion and not its magnitude. Given the direction, we can apply a small time step to update the object’s pose in the gripper contact frame. This small time step approximates the instantaneous object motion. Thus, we can predict the motion iteratively:
\begin{equation}
    \vec{q_{t+1}} = \vec{q_t} + \frac{\mat{A} \vec{w_f}}{||\mat{A} \vec{w_f}||}\delta t
    \label{eqn:motion_update}
\end{equation}
% \todo{This paragraph is super hard to understand, to the point where I am not able to edit it for clarity. We can either discuss it or you can have a crack at creating more clarity before I try again to edit.}
By employing sufficiently small time steps, we can approximate the trajectory without requiring time increments that scale linearly with the motion. To regulate the update, we instead use a unit-distance step. However, because the twist is normalized—without appropriately matching the scales of linear and angular velocities—the system exhibits faster motion when it is closer to pure rotation than when it is primarily translational. This limitation arises from relying solely on the limit surface (LS) model under quasi-static assumptions.

\begin{figure*}
    \centering
    \includegraphics[width=0.9\linewidth]{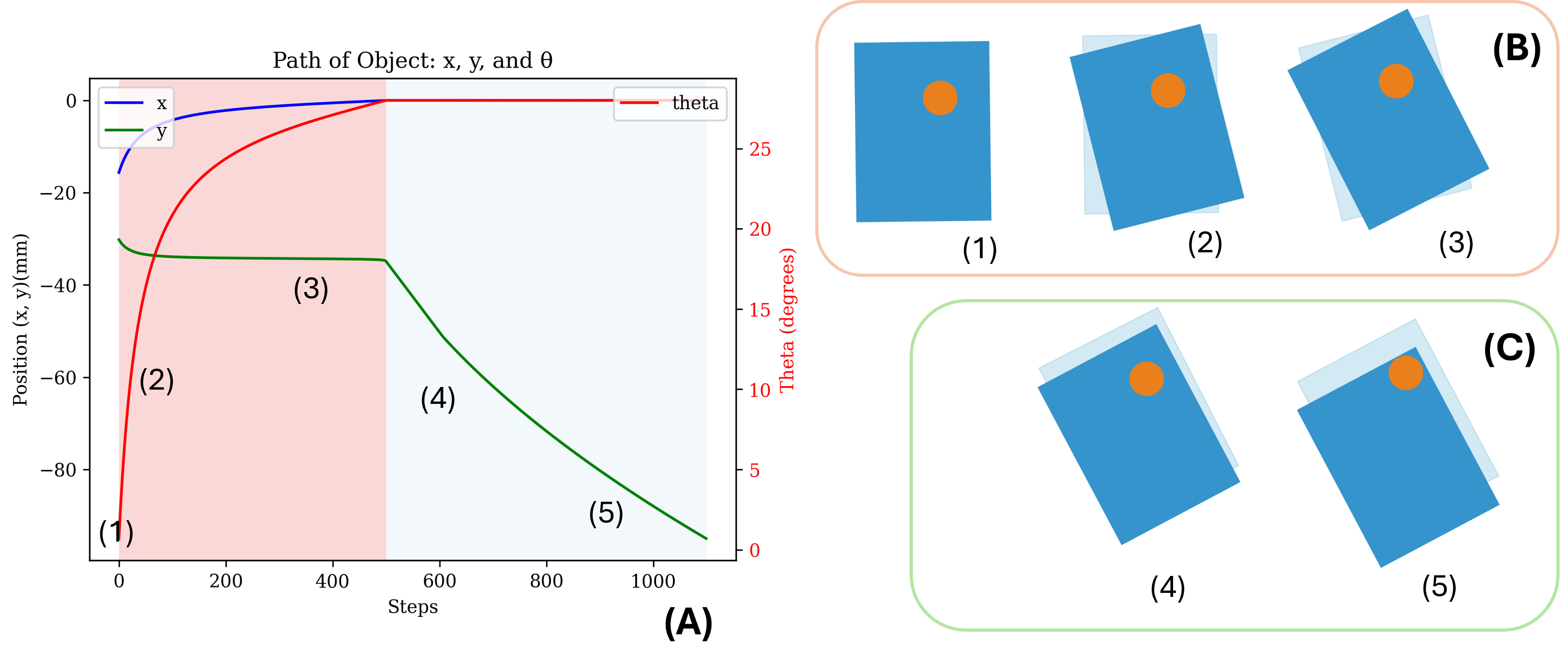}
    \caption{Example of object motion prediction using our motion model. \textbf{(A)} Predicted object path, highlighting two distinct motion types during sliding: near-rotational motion (pink region) and translational motion (light cyan region).\textbf{(B)} Visualization of near-rotational motion at steps (1), (2), and (3). \textbf{(C)} Visualization of translation motion at steps (4) and (5). Light blue indicates the previous pose, blue indicates the current pose, and the orange circle denotes the robot finger..}
    \label{fig:motion_prediction}
\end{figure*}

In Eq.~\ref{eqn:ft_balance}, enforcing force and torque balance effectively fixes the size of the limit surface, which reflects 
% \todo{What is frictional influence? Also, a clearer explanation is needed for why the limit surface size is fixed because of enforcing FT balance}
amount of friction needed to balance the object. A larger limit surface implies a greater frictional requirement to maintain equilibrium. We treat the scale of this limit surface intuitively as a scale factor for the twist, leading to a modified motion update equation:
\begin{equation}
    \begin{aligned}
        \vec{q_{t+1}} = \vec{q_t} + \frac{\mat{A} \vec{w_f}}{||\mat{A} \vec{w_f}||}\delta tk\\
        k = \frac{1}{\vec{w_f}^T\mat{A}\vec{w_f}}\
    \end{aligned}    
    \label{eqn:motion_update_modified}
\end{equation}
where $k$ is proportional to the limit surface size.
With this modified motion update method, we can obtain a more realistic motion prediction. 
One sample of in-hand object motion is plotted in Fig. \ref{fig:motion_prediction}.

\subsection{Vibration}
The motion model described above provides the foundation for understanding object reconfiguration under quasi-static assumption. Vibration is the means by which we are able to modify friction at the contact interface which in turn affects motion. Fingertip vibrations reduce the effective average normal force imparted to the object, thus reducing the effective friction force. This is because contact between the object and robot is continuously and very rapidly broken. 

Vibration has been extensively studied in tribology, with both theoretical studies and experimental results demonstrating that sliding friction decreases when vibrations are applied either tangentially \citep{gutowski2012effect} or normal \citep{hao2018effect} to the contact surface. The primary advantage of using vibration, rather than directly controlling the grasping force, lies in its ability to change state on the order of milliseconds. Conventional grippers, on the contrary, require more time to close and cannot achieve stable grasping forces in such a short interval. For our task, the ability to change rapidly is critical.

\section{Motion Planning for In-hand Reconfiguration} \label{sec:planning}
\begin{figure}[ht]
    \centering
    \includegraphics[width=0.7\linewidth]{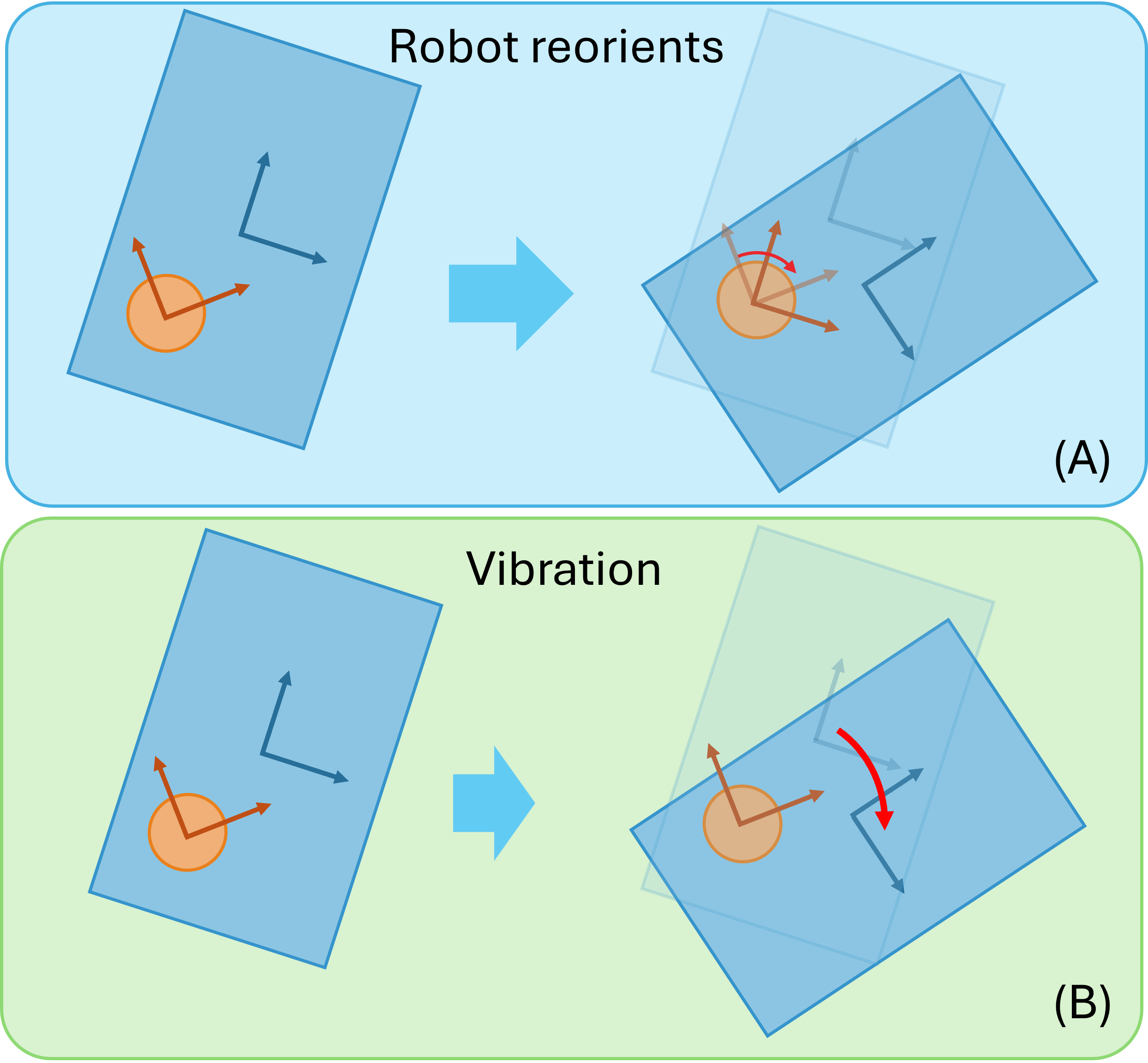}
    \caption{Two types of robot actions: (a) robot reorients the gripper without vibration, (b) robot stops moving (fixed gripper pose) and vibrates the fingertips allowing the object to slide.}
    \label{fig:robot_motions}
\end{figure}

In this section, we develop our proposed planning algorithm that guides the object to its target pose. Under the assumption that the object only slides when vibration is applied, we define two primitives, illustrated in Fig.~\ref{fig:robot_motions}:
\begin{enumerate}
    \item Reorient the gripper without vibration -- the relative pose between the object and the fingertip is fixed. However, by reorienting the gripper and object, we alter the gravitational torque on the gripper.
    \item Keep the gripper’s pose fixed and apply vibration --  we apply vibration to gradually change the relative pose through sliding, as described in the previous section.
\end{enumerate}

Primitive (a) affords the ability to predict the future object motions using Eq.~\ref{eqn:motion_update_modified} and primitive (b) realizes the motion. Since Eq.~\ref{eqn:motion_update_modified} does not provide a precise timestep, we use pose feedback to locate the object rather than execute in open-loop.

The key insight we use to derive our control strategy is obtained by carefully studying our model's predictions. As illustrated in Fig.~\ref{fig:motion_prediction}, under robot motion type (b), all object motions can effectively be grouped into two categories:
\begin{enumerate}
    \item Linear motion along the direction of gravity.
    \item Near-pure rotational motion.
\end{enumerate}

Linear motion along the direction of gravity occurs when the line connecting the object’s center of mass (CoM) and the contact patch’s center is parallel to gravity. This scenario has two distinct cases depending on whether the CoM lies above or below the contact center. If the CoM is above, the motion is sensitive to initial pose errors and can diverge to the second category. Conversely, if the CoM is below the contact center, the object slides stably along the gravity direction.

The object tends toward a near-pure rotational motion about the contact patch when the condition for the previous case is not met. In summary, the contact can either translate toward or away from the object’s center or rotate until the line connecting the object’s center of mass and the contact patch’s center aligns with gravity. This behavior makes it difficult to formulate MPC or other conventional control methods, since it is non-convex and requires a long horizon to predict a feasible path.

\begin{figure*}
    \centering
    \includegraphics[width=0.9\linewidth]{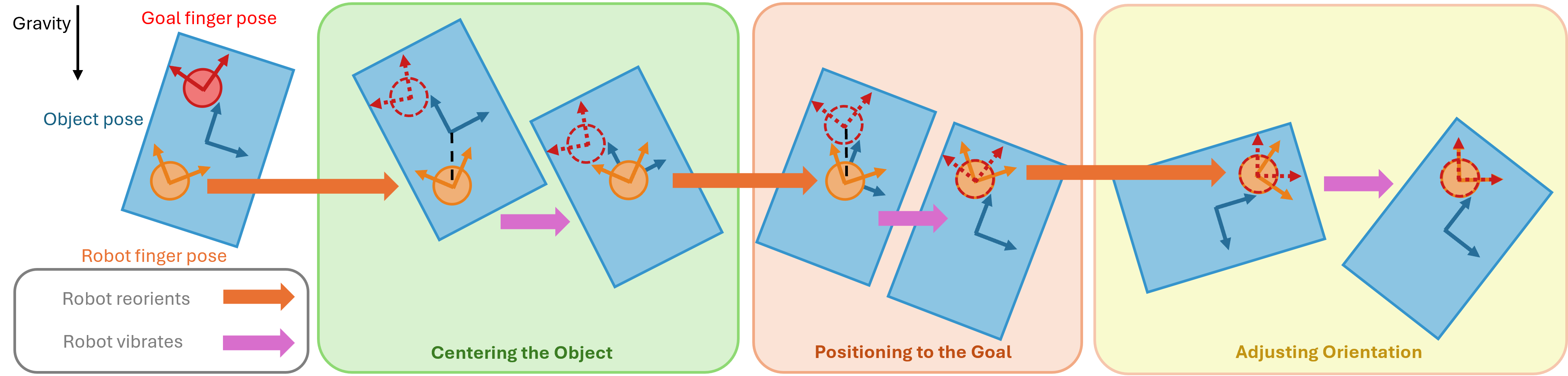}
    \caption{The step-by-step in-hand reconfiguration pipeline}
    \label{fig:subgoals}
\end{figure*}

Consequently, we propose a heuristic approach for generating a feasible path, informed by our understanding of the sliding motion behavior (see Fig.~\ref{fig:subgoals}). 
Based on the aforementioned properties, we observe that the object can be rotated at any pose as long as gravitational torque is present, and that it can only be translated toward or away from its center of mass. Our planning pipeline is outlined as follows:
\begin{enumerate}    
    \item \textbf{Centering the Object:} Repeat the following motion until it reaches within an error ball at center of gripper: $||x_e^w - x_o^w, y_e^w - y_o^w|| < r_{error}$: Rotate the gripper so that the object center is right above the gripper, then vibrate.
    \item \textbf{Positioning to the Goal:} Repeat the following motion until it reaches within an error ball at goal position: $||x_e^w - x_g^w, y_e^w - y_g^w|| < r_{error}$: Rotate the gripper so that the object center is directly below the goal pose, then vibrate.
    \item \textbf{Adjusting Orientation:} Repeat the following motion until the relative orientation reaches the goal: $||\theta_e^w -\theta_o^w -theta_g^o||<r_{\theta error}$: Rotate the gripper so that the object center is not right above or below gripper, then vibrate.
\end{enumerate}

At the end of the first or second stage, when the distance between the finger and the object’s center approaches zero, the system enters a singularity that destabilizes the finger’s orientation. To mitigate this, we define an error region with radius $r_{error}$. Once the finger is within this error region, we proceed to the next stage. If $r_{error}$ is too large, the finger remains far from the object’s center, potentially causing excessive gravitational torque and motion failure in the next stage. Conversely, if 
$r_{error}$ is too small, the system may approach singularity and fail. In our setup, we chose to use 5 [mm] as the radius $r_{error}$.

In practice, the pressure center of the patch contact does not always coincide with the geometric center of the contact due to friction throughout the gripper system. Therefore, we estimate the pressure center to execute a feasible path. Additionally, because the system is unstable in the first stage, requiring pose feedback for stable execution, we incorporate a PI controller at each step to compensate for system errors or adjust pose estimates. Our system operates through discretized motion, it behaves as a first-order system, where the PI controller ensures stable control. As outlined in Algorithm~\ref{alg:vibration_reconfiguration}, each vibration step includes an observation step to update the estimated object state, ensuring a feasible and controlled motion throughout the reconfiguration process.
% \todo{pipeline figure / or a pseudo code}. 
We validate our planning and control algorithm in the results section.

\begin{algorithm}
\caption{Vibration-Based In-Hand Reconfiguration}
\label{alg:vibration_reconfiguration}
\begin{algorithmic}
\State \textbf{Input:} Initial object pose $(x_o, y_o, \theta_o)$, goal pose $(x_g, y_g, \theta_g)$
\State \textbf{Output:} Object at $(x_g, y_g, \theta_g)$

\State \textbf{Note:} All steps incorporate feedback from the estimated contact center for correction.

\State \textbf{Step 1: Centering the Object}
\While{$||x_e^w - x_o^w, y_e^w - y_o^w|| \geq r_{error}$}
    \State Rotate gripper to align object center above contact center
    \State Apply vibration
    \State Observe object state and update estimate
\EndWhile

\State \textbf{Step 2: Moving to Goal}
\While{$||x_e^w - x_g^w, y_e^w - y_g^w|| \geq r_{error}$}
    \State Rotate gripper so object center aligns with goal
    \State Apply vibration
    \State Observe object state and update estimate
\EndWhile

\State \textbf{Step 3: Adjusting Orientation}
\While{$||\theta_e^w - \theta_o^w - \theta_g^o|| \geq r_{\theta error}$}
    \State Rotate gripper to apply gravitational torque
    \State Apply vibration
    \State Observe object state and update estimate
\EndWhile

\State \textbf{Return:} Object positioned and oriented at goal.

\end{algorithmic}
\end{algorithm}

\section{Experiments}
In our experiments, we evaluate both our motion model and our planning and control pipeline for free-space in-hand reconfiguration. All experiments are conducted in real-world settings, due to the challenges of accurately simulating vibration.

\subsection{Experiment setup}
\begin{figure}
    \centering
    \includegraphics[width=1.0\linewidth]{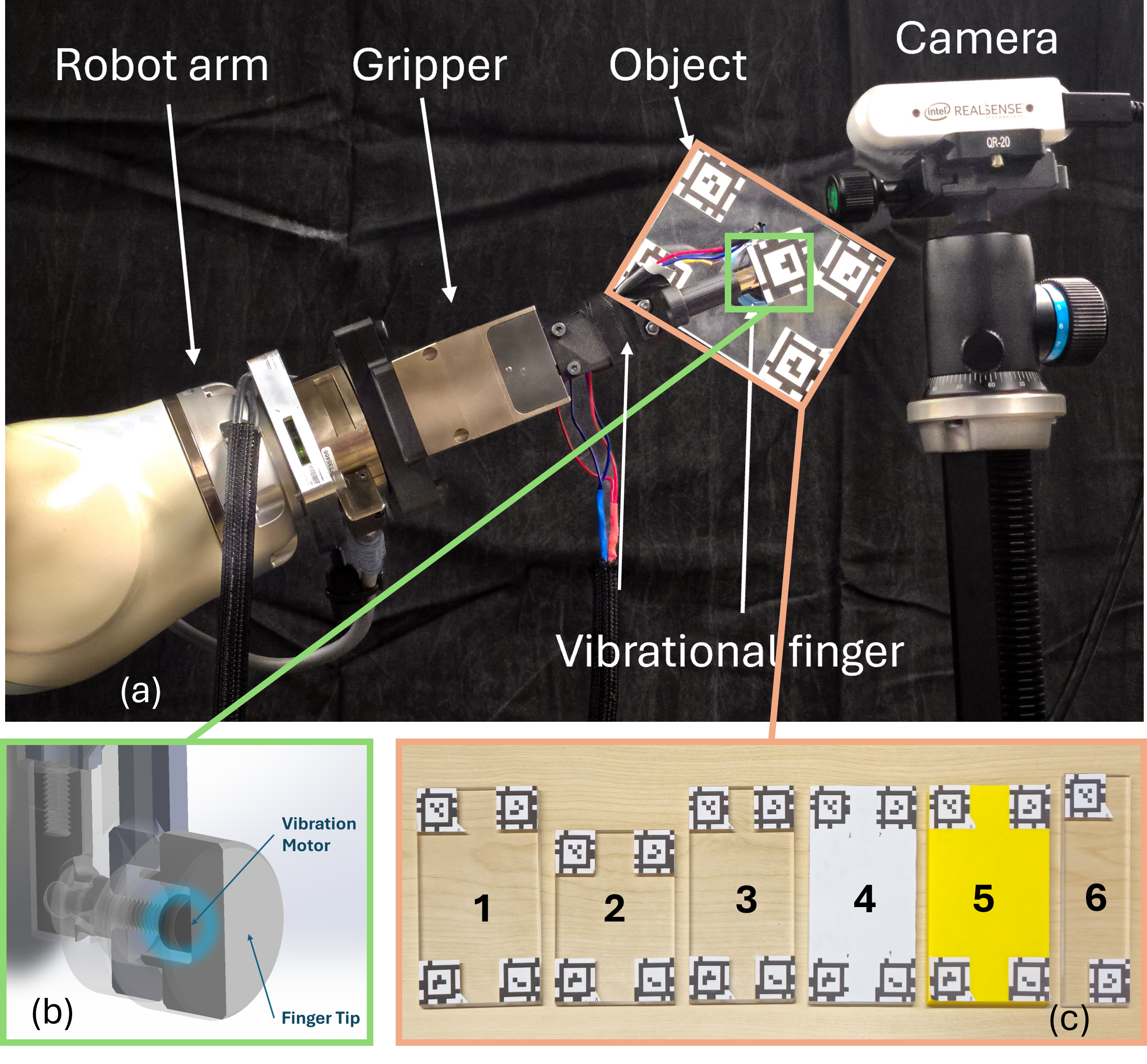}
    \caption{Experimental setup: (a) Overview of the experimental setup; (b) Detailed view of the vibrational fingertip featuring an embedded vibration motor; (c) Objects equipped with AprilTags for localization.}
    \vspace{-0.5cm}
    \label{fig:setup}
\end{figure}

For our real-world experiments, we mount the vibration-based fingers on a WSG-50 gripper, which is attached to a Kuka iiwa Med R820 robot arm, as shown in Fig.\ref{fig:setup}(a). The finger attachments are PLA 3D-printed and feature a hollow chamber to house the vibration motors (Fig.\ref{fig:setup}(b)). Each fingertip incorporates a 10 [mm] diameter DC vibration motor (Model: \textit{DXD-B1030X50-3CW-0.3PS coin motor}, Zard Zoop). The contact surface of the fingertip is 30 [mm] in diameter, and to enhance compliance and achieve a more uniform pressure distribution, we cover the surface with paper tape. Two vibration motors, one on each side, share the same power supply and are controlled by a Raspberry Pi 3B running ROS. The motors operate at a constant voltage of 4.0 [V] using on-off switching via a relay. This design is simple, reproducible, and adaptable for other gripper platforms.

We define a vertical working plane for the robot so that the gripper and object operate in an SE(2) plane, chosen to maximize the robot’s workspace. An Intel RealSense D415 camera is fixed to observe this plane, providing real-time object pose estimation. 
All experimental objects are planar, including six shapes with different geometries, surface materials, and mass distributions. Specifically, we use five acrylic plates cut with a 
$\text{CO}_2$ laser, and one 3D-printed PLA plate. We also cover one acrylic plate with paper on both sides to provide different surface property. Each object is equipped with AprilTags for pose estimation (Fig.~\ref{fig:setup}(c)). Object weights from 53 [g] to 90 [g]. Details about the objects are listed in Table. \ref{tab:planner_error}.
% All experimental objects are acrylic plates cut with a $\text{CO}_2$ laser, each equipped with four AprilTags for pose estimation Fig.~\ref{fig:setup}(c). Their masses are $m=\{71, 79, 90\}$[g].

% We also attach an AprilTag to the back of the robot gripper finger to track the finger’s pose relative to the object center, which provides higher accuracy than the robot’s internal sensors.

\subsection{Motion Prediction Evaluation}
To evaluate the in-hand sliding motion model, we compare the sliding path of the same scene in real-world experiments and motion predicted using our model. For each run, we randomly choose the object pose, and fix the gripper pose, then take a single vibration action, measuring the resulting relative pose between the robot and object. We replicate the same scene setup with the same contact patch radius (15 [mm]) and initial object pose using our model for evaluation. We compute the RMSE error between the measured and predicted object pose with respect to the robot. We collected 1564 vibration actions.
% \todo{Maybe not the last, as we don't know what is the last step.}

\subsection{Real-world Experiment Procedure}
To evaluate the performance of our motion planning and control pipeline, we first set the robot arm to a preset home pose within the working plane, then open the gripper and put the object between fingers. Next, we close the gripper and apply a preset grasping force. Once the object is secured,  we start our motion pipeline and record object pose and gripper pose during the motion execution. 
We executed 10 motions for each object, and record the path for further evaluation. The root mean square error (RMSE) between the measured and goal object poses, computed with respect to the robot frame, is used as the performance metric.

For reproducibility, we specify the following details of the experimental setup:
\begin{itemize}
    \item \textbf{Grasping force:} Empirically set to approximately $5$ [N], sufficient to prevent the object from dropping in the absence of vibration while allowing controlled slippage during actuation.
    \item \textbf{Pressure center estimation:} Assumed to be located at the geometric center of the fingertip contact surface.
    \item \textbf{Initial and goal poses:} Goal poses are randomly sampled within $50$ [mm] translation and $\pm 60^\circ$ rotation relative to the initial pose. To ensure feasibility within the robot’s joint limits, we discard any sampled goals that require robot end-effector finger orientation outside the range of $[-1.2, 1.5]$ [rad].
\end{itemize}

\section{Experiment Results}
% \textcolor{orange}{On-going work}
% In this section, we discuss the results of in-hand sliding.

\subsection{Motion model evaluation}
Table. \ref{tab:model_error} shows the quantitative results of model evaluations as a function of the timestep chosen for our model. For all vibration actions, the RMSE of goal pose is 1.849 [mm], with 2.189 [degree]. We also note that the pose obtained using the RealSense and AprilTags system exhibits noise greater than 1 [mm]. These results are obtained by setting the robot to a random configuration and taking a single vibration activation action. The loss is computed between the predicted and measured object pose with respect to the end effector. We note that the prediction error is sufficiently small to suggest that closed-loop control will be effective; however, there is a sufficient discrepancy in predictions that would render open-loop execution impractical.

% \todo{Placeholder for the table. Update values as needed.}
\begin{table}[ht]
\centering
\caption{Path error of proposed model}
\label{tab:model_error}
\begin{tabular}{c | c c}
\hline
\textbf{} & \textbf{RMSE Position (mm)} & \textbf{RMSE Orientation (degree)} \\ \hline
\textbf{Value} & 1.849 & 2.189 \\ \hline
\end{tabular}
\end{table}

\begin{figure}
    \centering
    \includegraphics[width=0.85\linewidth]{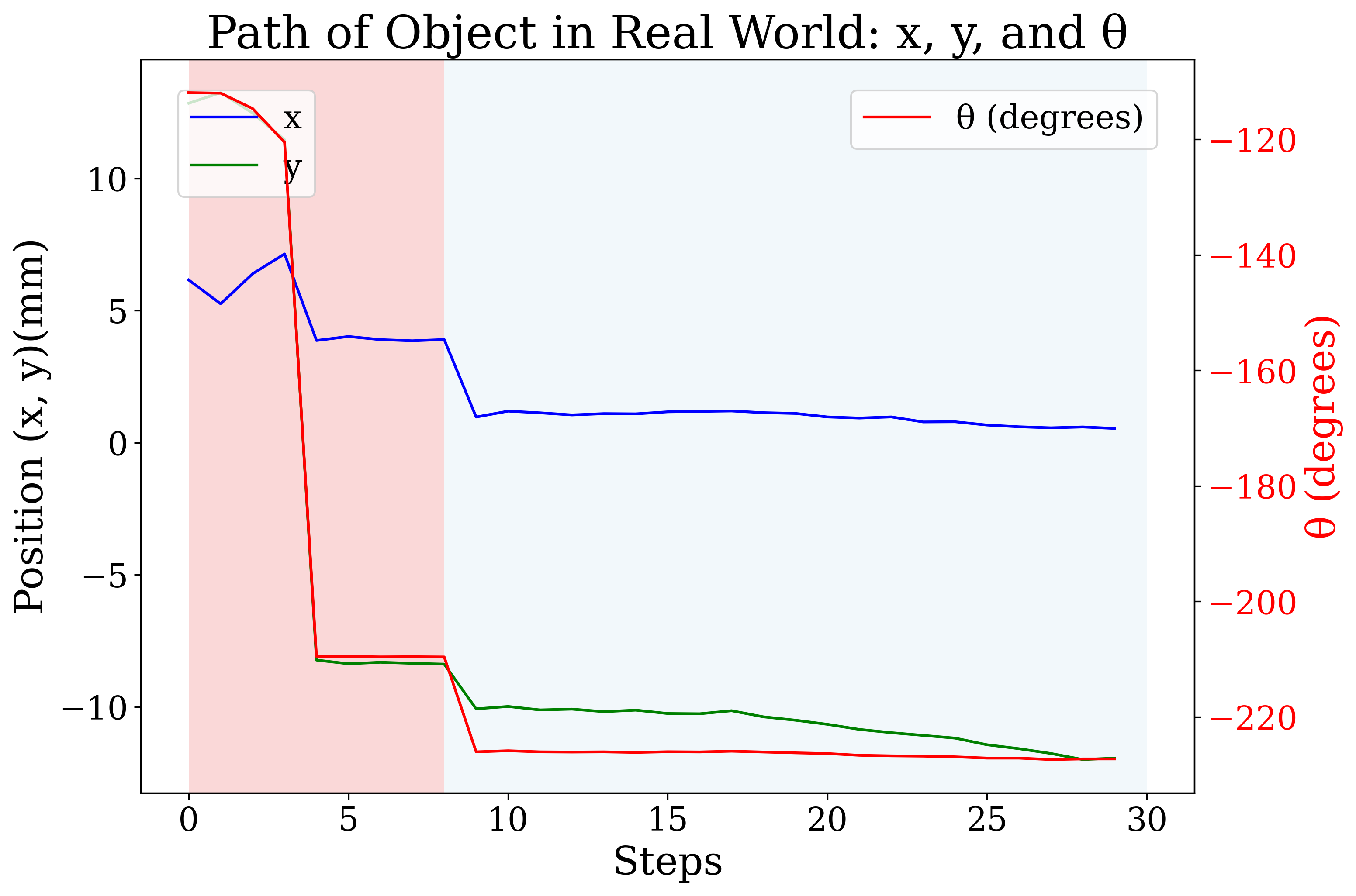}
    \caption{Visualization of paths from real-world experiments. From the path, we can devide the motion into two kinds: the pink and blue background represent near-rotation motion and translational motion.}
    \label{fig:model_path_vis}
\end{figure}
To provide a more intuitive understanding, we also show one whole path of object motion with multiple vibration actions in real-world in Fig. \ref{fig:model_path_vis}.  We find that there are 2 stages of motion, as described in Section \ref{sec:planning}, the first resembles near-rotation motion, while the second resembles pure translation. We also notice that most error comes from the first rotational stage. This is mostly explained by unpredictable variation in both the contact pressure center and the equivalent radius of the contact patch.

\subsection{In-hand reconfiguration evaluation}
In this section, we evaluate our in-hand reconfiguration planner over a variety of paths. We report the root mean square error (RMSE) of the object’s position and orientation at the end of each motion, along with the relative position error, defined as the RMSE normalized by the object’s max length.

% \todo{Placeholder for the table. Update values as needed.}

\begin{table*}  % Use [b] for bottom placement
\centering
\vspace{-0.3cm}
\begin{tabular}{ c c c c c c c}
% \hline
\textbf{Object \#}& \textbf{Surface Material} & \textbf{Size (mm)} & \textbf{Weight (g)} & \textbf{RMSE Pos. (mm)} & \textbf{Rel. Error (\%)} & \textbf{RMSE Orient. (deg)} \\
\hline
1 & Acrylic & 90$\times$150 & 90 & 6.59 & 4.4\% & 1.75 \\
2 & Acrylic & 90$\times$120 & 71 & 4.36 & 3.6\% & 0.89 \\
3 & Acrylic & 80$\times$150 & 79 & 5.79 & 3.9\% & 1.11 \\
4 & Paper & 80$\times$150  & 85 & 5.33 & 3.6\% & 1.00 \\
5 & PLA   & 90$\times$150  & 92 & 6.36 & 4.2\% & 0.99 \\
6 & Acrylic & 50$\times$160  & 53 & 7.04 & 4.4\% & 1.53 \\ \hline
\textbf{Average} & ---  & --- & --- & 5.91 & --- & 1.21 \\
\hline
\end{tabular}
\caption{Final pose error statistics over 10 trials for each object. }
\label{tab:planner_error}
\end{table*}

\begin{figure}[t]
    \centering
    \includegraphics[width=1.0\linewidth]{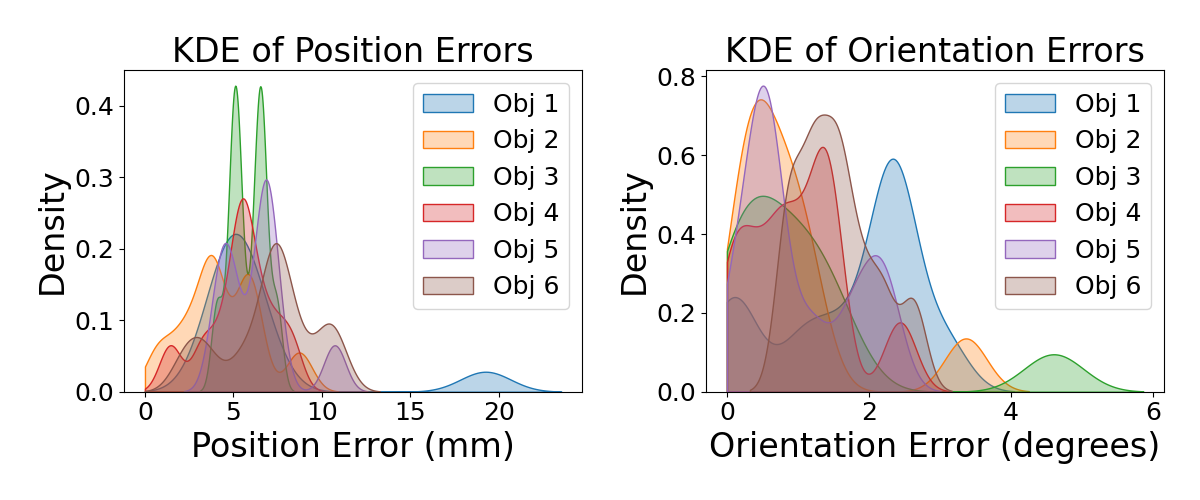}
    \caption{Final pose error distribution of each object, and the summary quantitative results of this figure are represented in Table. ~\ref{tab:planner_error}.}
    \label{fig:pose_error_distribution}
\end{figure}

For all 60 tested paths across six different objects, the root mean square error (RMSE) at the final pose is \(5.91\) [mm] in position and \(1.21^\circ\) in orientation. To better contextualize these results, we report detailed error statistics in Table. ~\ref{tab:planner_error} and visualize the error distribution in Fig.~\ref{fig:pose_error_distribution}. The observed position errors correspond to approximately \(3\%\)–\(5\%\) of the object dimensions. We note that the combination of the Intel RealSense D415 camera and AprilTag-based localization introduces a baseline measurement noise greater than \(1\) [mm], which contributes to the reported errors.

Our evaluation demonstrates that the proposed method reliably achieves accurate in-hand reconfiguration across objects with varying shapes, surface materials, and mass distributions. As summarized in Table. ~\ref{tab:planner_error}, the system successfully reconfigured all six tested objects. Notably, Object \#5, which is 3D-printed with diagonal infill paths, presents non-homogeneous frictional properties due to its textured surface. Under vibration, this object exhibits motion bias along the direction of the texture combined with gravity, rather than purely along the gravity direction. Despite this added complexity, our method consistently achieved a final position error of less than \(7\) [mm], demonstrating robustness to friction anisotropy and material inconsistencies.

These results suggest that the proposed approach is effective and robust under real-world conditions, including variations in object geometry, surface texture, and mass distribution.

\begin{figure*}
    \centering
    \includegraphics[width=0.9\linewidth]{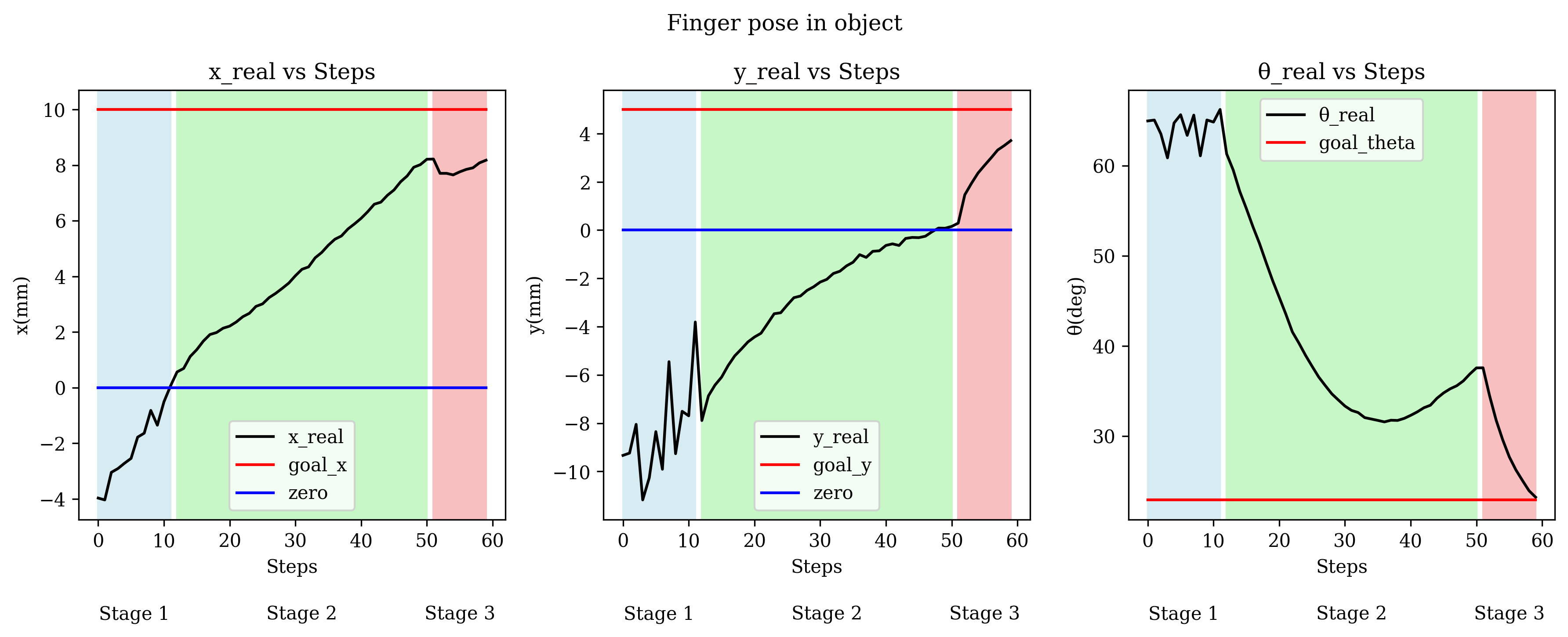}
    \caption{Finger motion in object frame, with goal pose [10, 5, 22.9]. Background color light blue, light green and light pink represent 3 stages of our planning. The red lines represent goal pose, the blue lines represent object center position, and the black curve represent finger motion in object frame. }
    \label{fig:path_sample}
\end{figure*}

\begin{figure*}
    \centering
    \includegraphics[width=0.95\linewidth]{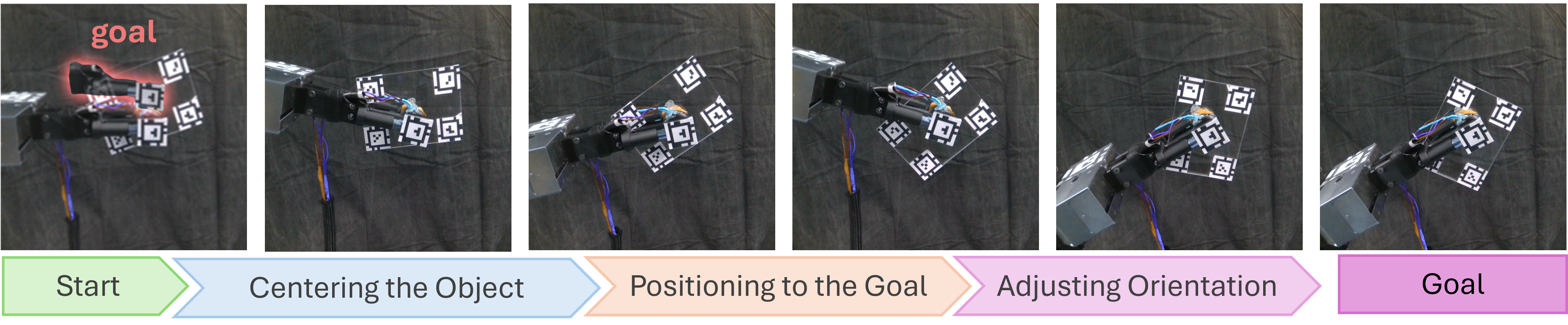}
    \caption{Visualization of robot and object motions to do in-hand reconfiguration. }
    \label{fig:path_sample_vis}
\end{figure*}
We also visualize several paths in Fig. \ref{fig:path_sample} to provide an illustration of the traversed path. We notice behavior differences in different stages. In the first stage, as the whole system is not stable, we observe finger orientation oscillations, partly due to limited control over the pressure center compensation at the beginning of the path. As the finger moves closer to object pose, the system gets closer to singularity, which results in finger oscillates once more. 

During the second stage, the system is in a more stable configuration, thus the finger orientation is more stable. Despite this relative stability, we observe some relative rotation between the finger and object due to unpredictable variations in the effective contact patch and pressure distribution (discussed in Sec.~\ref{sec:planning}). To address this relative orientation mismatch, we perform the final orientation adjustment, (as shown in Fig. \ref{fig:path_sample}). These observations further motivate our strategy to rotate the object after translation, instead of rotate first, then translate to goal pose. We also notice that during the \textbf{Adjusting Orientation} stage, the position accuracy drops due to the slight translation from near-rotation motion.
Fig. \ref{fig:path_sample_vis} shows a visualization of robot motions to complete the in-hand reconfiguration task via vibration.

% The in-hand manipulation results are listed in Table. \ref{}, 

\section{Discussion \& Limitations}
In this paper, we presented a vibration-based parallel gripper and motion planning and control algorithm for in-hand sliding in free space, to drive a planar object to arbitrary in-hand pose. Our goal is to enable more dexterity than that which is possible with conventional parallel jaw grippers. We utilize the physical phenomenon of slippage from vibration into a active control input. We also develop the friction and motion model for in-hand sliding in free space, where gravity is the only external force. With this motion model, we propose a heuristic based subgoal generation method and closed-loop control algorithm to reconfigure an object in hand. In real-world experiments, we show the effectiveness and robustness of our methods to reconfigure objects.

Perhaps the most limiting factor of our approach is that the objects can only be repositioned on parallel surfaces aligned with the grasp. This is the same limitation as any other method depending on in hand relative motion realized through the fingers \citep{chavan2020planar,hou2018fast}. A second limitation is that our method is formulated strictly in the gravity plane. By making use of planes at an angle to gravity, it is conceivable that more precise or diverse strategies can be found. Recent work such as \citep{nahum2022robotic, binyamin2024vibration}, have specific designs that have explored more tilted plane or even horizontal plane.

In terms of actuation, we treated vibration only as a simple friction changing source, and we assume that the change is isotropic and homogeneous across the surface of the object. Exploring directional effects or variation in vibration intensity could also improve and broaden the set of motions available to plan over (e.g., more fine grained control over rotations and translations). A secondary consideration is frequency, where lower frequency changes can potentially allow for more controlled but slower motion.

While we use vibration exclusively as a mode of actuation, it is plausible to use the feedback from vibration to make inferences about the contact mode (slipping vs sticking) or the grasped object. This is a potentially exciting avenue of exploration.

\bibliographystyle{plainnat}
\bibliography{references.bib}

\begin{thebibliography}{44}
\providecommand{\natexlab}[1]{#1}
\providecommand{\url}[1]{\texttt{#1}}
\expandafter\ifx\csname urlstyle\endcsname\relax
  \providecommand{\doi}[1]{doi: #1}\else
  \providecommand{\doi}{doi: \begingroup \urlstyle{rm}\Url}\fi

\bibitem[Chavan-Dafle et~al.(2020)Chavan-Dafle, Holladay, and Rodriguez]{chavan2020planar}
Nikhil Chavan-Dafle, Rachel Holladay, and Alberto Rodriguez.
\newblock Planar in-hand manipulation via motion cones.
\newblock \emph{The International Journal of Robotics Research}, 39\penalty0 (2-3):\penalty0 163--182, 2020.

\bibitem[Nahum and Sintov(2022)]{nahum2022robotic}
Noam Nahum and Avishai Sintov.
\newblock Robotic manipulation of thin objects within off-the-shelf parallel grippers with a vibration finger.
\newblock \emph{Mechanism and Machine Theory}, 177:\penalty0 105032, 2022.

\bibitem[Binyamin et~al.(2024)Binyamin, Shapira, Nahum, and Sintov]{binyamin2024vibration}
Oron Binyamin, Guy Shapira, Noam Nahum, and Avishai Sintov.
\newblock Vibration-based full state in-hand manipulation of thin objects.
\newblock \emph{arXiv preprint arXiv:2412.14899}, 2024.

\bibitem[Pfanne et~al.(2020)Pfanne, Chalon, Stulp, Ritter, and Albu-Sch{\"a}ffer]{pfanne2020object}
Martin Pfanne, Maxime Chalon, Freek Stulp, Helge Ritter, and Alin Albu-Sch{\"a}ffer.
\newblock Object-level impedance control for dexterous in-hand manipulation.
\newblock \emph{IEEE Robotics and Automation Letters}, 5\penalty0 (2):\penalty0 2987--2994, 2020.

\bibitem[Jiang et~al.(2024)Jiang, Yu, Zhu, Tomizuka, and Li]{jiang2024contact}
Yongpeng Jiang, Mingrui Yu, Xinghao Zhu, Masayoshi Tomizuka, and Xiang Li.
\newblock Contact-implicit model predictive control for dexterous in-hand manipulation: A long-horizon and robust approach.
\newblock \emph{arXiv preprint arXiv:2402.18897}, 2024.

\bibitem[Luo et~al.(2024)Luo, Li, Wang, Duan, Wei, and Sun]{luo2024progressive}
Yongkang Luo, Wanyi Li, Peng Wang, Haonan Duan, Wei Wei, and Jia Sun.
\newblock Progressive transfer learning for dexterous in-hand manipulation with multi-fingered anthropomorphic hand.
\newblock \emph{IEEE Transactions on Cognitive and Developmental Systems}, 2024.

\bibitem[Chen et~al.(2022)Chen, Xu, and Agrawal]{chen2022system}
Tao Chen, Jie Xu, and Pulkit Agrawal.
\newblock A system for general in-hand object re-orientation.
\newblock In \emph{Conference on Robot Learning}, pages 297--307. PMLR, 2022.

\bibitem[Zarrin et~al.(2023)Zarrin, Jitosho, and Yamane]{zarrin2023hybrid}
Rana~Soltani Zarrin, Rianna Jitosho, and Katsu Yamane.
\newblock Hybrid learning-and model-based planning and control of in-hand manipulation.
\newblock In \emph{2023 IEEE/RSJ International Conference on Intelligent Robots and Systems (IROS)}, pages 8720--8726. IEEE, 2023.

\bibitem[Veiga et~al.(2020)Veiga, Akrour, and Peters]{veiga2020hierarchical}
Filipe Veiga, Riad Akrour, and Jan Peters.
\newblock Hierarchical tactile-based control decomposition of dexterous in-hand manipulation tasks.
\newblock \emph{Frontiers in Robotics and AI}, 7:\penalty0 521448, 2020.

\bibitem[Arunachalam et~al.(2023)Arunachalam, Silwal, Evans, and Pinto]{arunachalam2023dexterous}
Sridhar~Pandian Arunachalam, Sneha Silwal, Ben Evans, and Lerrel Pinto.
\newblock Dexterous imitation made easy: A learning-based framework for efficient dexterous manipulation.
\newblock In \emph{2023 ieee international conference on robotics and automation (icra)}, pages 5954--5961. IEEE, 2023.

\bibitem[Solak and Jamone(2019)]{solak2019learning}
Gokhan Solak and Lorenzo Jamone.
\newblock Learning by demonstration and robust control of dexterous in-hand robotic manipulation skills.
\newblock In \emph{2019 ieee/rsj international conference on intelligent robots and systems (iros)}, pages 8246--8251. IEEE, 2019.

\bibitem[Hammoud et~al.(2024)Hammoud, Belcamino, Huet, Carfi, Khoramshahi, Perdereau, and Mastrogiovanni]{hammoud2024robotic}
Ali Hammoud, Valerio Belcamino, Quentin Huet, Alessandro Carfi, Mahdi Khoramshahi, Veronique Perdereau, and Fulvio Mastrogiovanni.
\newblock Robotic in-hand manipulation with relaxed optimization.
\newblock \emph{arXiv preprint arXiv:2406.04950}, 2024.

\bibitem[Karayiannidis et~al.(2016)Karayiannidis, Smith, Kragic, et~al.]{karayiannidis2016adaptive}
Yiannis Karayiannidis, Christian Smith, Danica Kragic, et~al.
\newblock Adaptive control for pivoting with visual and tactile feedback.
\newblock In \emph{2016 IEEE International Conference on Robotics and Automation (ICRA)}, pages 399--406. IEEE, 2016.

\bibitem[Hou et~al.(2018)Hou, Jia, and Mason]{hou2018fast}
Yifan Hou, Zhenzhong Jia, and Matthew~T Mason.
\newblock Fast planning for 3d any-pose-reorienting using pivoting.
\newblock In \emph{2018 IEEE International Conference on Robotics and Automation (ICRA)}, pages 1631--1638. IEEE, 2018.

\bibitem[Ma and Dollar(2016)]{ma2016hand}
Raymond~R Ma and Aaron~M Dollar.
\newblock In-hand manipulation primitives for a minimal, underactuated gripper with active surfaces.
\newblock In \emph{International Design Engineering Technical Conferences and Computers and Information in Engineering Conference}, volume 50152, page V05AT07A072. American Society of Mechanical Engineers, 2016.

\bibitem[St{\"u}ber et~al.(2020)St{\"u}ber, Zito, and Stolkin]{stuber2020let}
Jochen St{\"u}ber, Claudio Zito, and Rustam Stolkin.
\newblock \href{https://www.frontiersin.org/articles/10.3389/frobt.2020.00008/full}{Let's push things forward: A survey on robot pushing}.
\newblock \emph{Frontiers in Robotics and AI}, page~8, 2020.

\bibitem[Lynch and Mason(1996)]{lynch1996stable}
Kevin~M Lynch and Matthew~T Mason.
\newblock \href{https://journals.sagepub.com/doi/10.1177/027836499601500602}{Stable pushing: Mechanics, controllability, and planning}.
\newblock \emph{The international journal of robotics research}, 15\penalty0 (6):\penalty0 533--556, 1996.

\bibitem[Kao and Cutkosky(1992)]{kao1992quasistatic}
Imin Kao and Mark~R Cutkosky.
\newblock \href{https://ieeexplore.ieee.org/document/351048}{Quasistatic manipulation with compliance and sliding}.
\newblock \emph{The International journal of robotics research}, 11\penalty0 (1):\penalty0 20--40, 1992.

\bibitem[Kao and Cutkosky(1993)]{kao1993comparison}
Imin Kao and Mark~R Cutkosky.
\newblock \href{https://journals.sagepub.com/doi/10.1177/027836499301200602}{Comparison of theoretical and experimental force/motion trajectories for dextrous manipulation with sliding}.
\newblock \emph{The International journal of robotics research}, 12\penalty0 (6):\penalty0 529--534, 1993.

\bibitem[Xue and Kao(1994)]{xue1994dexterous}
Ying Xue and Imin Kao.
\newblock \href{}{Dexterous sliding manipulating using soft fingertips}.
\newblock In \emph{Proceedings of the 1994 IEEE International Conference on Robotics and Automation}, pages 3397--3402. IEEE, 1994.

\bibitem[Ghazaei~Ardakani et~al.(2020)Ghazaei~Ardakani, Bimbo, and Prattichizzo]{ghazaei2020quasi}
M~Mahdi Ghazaei~Ardakani, Joao Bimbo, and Domenico Prattichizzo.
\newblock \href{https://journals.sagepub.com/doi/full/10.1177/0278364920929082}{Quasi-static analysis of planar sliding using friction patches}.
\newblock \emph{The International Journal of Robotics Research}, 39\penalty0 (14):\penalty0 1775--1795, 2020.

\bibitem[Fakhari et~al.(2019)Fakhari, Kao, and Keshmiri]{fakhari2019modeling}
Amin Fakhari, Imin Kao, and Mehdi Keshmiri.
\newblock \href{https://robomechjournal.springeropen.com/articles/10.1186/s40648-019-0143-0}{Modeling and control of planar slippage in object manipulation using robotic soft fingers}.
\newblock \emph{ROBOMECH Journal}, 6\penalty0 (1):\penalty0 1--11, 2019.

\bibitem[Sampaziotis and Doulgeri(2022)]{9811686}
Savvas Sampaziotis and Zoe Doulgeri.
\newblock \href{https://ieeexplore.ieee.org/document/9811686}{A model free robot control method for dragging an object on a planar surface by applying top contact forces.}
\newblock In \emph{2022 International Conference on Robotics and Automation (ICRA)}, pages 8361--8367, 2022.
\newblock \doi{10.1109/ICRA46639.2022.9811686}.

\bibitem[Yi and Fazeli(2023)]{yi2023precise}
Xili Yi and Nima Fazeli.
\newblock \href{https://www.roboticsproceedings.org/rss19/p045.pdf}{Precise Object Sliding with Top Contact via Asymmetric Dual Limit Surfaces}.
\newblock \emph{Robotics: Science and Systems}, 2023.

\bibitem[Yi et~al.(2024)Yi, Dang, and Fazeli]{yi2024dual}
Xili Yi, An~Dang, and Nima Fazeli.
\newblock \href{https://link.springer.com/article/10.1007/s10514-024-10173-5}{Dual asymmetric limit surfaces and their applications to planar manipulation}.
\newblock \emph{Autonomous Robots}, 48\penalty0 (7):\penalty0 21, 2024.

\bibitem[Liu et~al.(2020)Liu, Ni, Wang, and Zhao]{liu2020analytical}
Weili Liu, Hongjian Ni, Peng Wang, and Bo~Zhao.
\newblock Analytical investigation of the friction reduction performance of longitudinal vibration based on the modified elastoplastic contact model.
\newblock \emph{Tribology International}, 146:\penalty0 106237, 2020.

\bibitem[Holl et~al.(2018)Holl, Meindlhumer, Simader, Schn{\"u}rer, and Brandl]{holl2018experimental}
Helmut~J Holl, Martin Meindlhumer, Viktoria Simader, Dominik Schn{\"u}rer, and Andreas Brandl.
\newblock Experimental investigation of friction reduction: by superimposed vibrations.
\newblock \emph{Materials Today: Proceedings}, 5\penalty0 (13):\penalty0 26615--26621, 2018.

\bibitem[Winkler(1978)]{winkler1978analysing}
G~Winkler.
\newblock Analysing the vibrating conveyor.
\newblock \emph{International Journal of Mechanical Sciences}, 20\penalty0 (9):\penalty0 561--570, 1978.

\bibitem[Gao et~al.(1994)Gao, Kuhlmann-Wilsdorf, and Makel]{gao1994dynamic}
Chao Gao, Doris Kuhlmann-Wilsdorf, and David~D Makel.
\newblock The dynamic analysis of stick-slip motion.
\newblock \emph{Wear}, 173\penalty0 (1-2):\penalty0 1--12, 1994.

\bibitem[Rubenstein et~al.(2014{\natexlab{a}})Rubenstein, Ahler, Hoff, Cabrera, and Nagpal]{rubenstein2014kilobot}
Michael Rubenstein, Christian Ahler, Nick Hoff, Adrian Cabrera, and Radhika Nagpal.
\newblock Kilobot: A low cost robot with scalable operations designed for collective behaviors.
\newblock \emph{Robotics and Autonomous Systems}, 62\penalty0 (7):\penalty0 966--975, 2014{\natexlab{a}}.

\bibitem[Rubenstein et~al.(2014{\natexlab{b}})Rubenstein, Cornejo, and Nagpal]{rubenstein2014programmable}
Michael Rubenstein, Alejandro Cornejo, and Radhika Nagpal.
\newblock Programmable self-assembly in a thousand-robot swarm.
\newblock \emph{Science}, 345\penalty0 (6198):\penalty0 795--799, 2014{\natexlab{b}}.

\bibitem[Maruo et~al.(2022)Maruo, Shibata, and Higashimori]{maruo2022dynamic}
Akihiro Maruo, Akihide Shibata, and Mitsuru Higashimori.
\newblock Dynamic underactuated manipulator using a flexible body with a structural anisotropy.
\newblock In \emph{2022 International Conference on Robotics and Automation (ICRA)}, pages 7117--7123. IEEE, 2022.

\bibitem[Yako et~al.(2024)Yako, Nowak, Yuan, and Salisbury]{yako2024vertical}
CL~Yako, J{\'e}r{\^o}me Nowak, Shenli Yuan, and Kenneth Salisbury.
\newblock Vertical vibratory transport of grasped parts using impacts.
\newblock In \emph{2024 IEEE International Conference on Robotics and Automation (ICRA)}, pages 1950--1956. IEEE, 2024.

\bibitem[Howe et~al.(1988)Howe, Kao, and Cutkosky]{howe1988sliding}
Robert~D Howe, Imin Kao, and Mark~R Cutkosky.
\newblock \href{https://ieeexplore.ieee.org/document/12032}{The sliding of robot fingers under combined torsion and shear loading}.
\newblock In \emph{Proceedings. 1988 IEEE International Conference on Robotics and Automation}, pages 103--105. IEEE, 1988.

\bibitem[Goyal(PhD Thesis, Cornell University Ithaca, NY, 1989)]{goyal1989planar}
Suresh Goyal.
\newblock \href{http://ruina.tam.cornell.edu/research/topics/friction_and_fracture/GoyalPhDThesis.pdf}{Planar sliding of a rigid body with dry friction: limit surfaces and dynamics of motion}.
\newblock PhD Thesis, Cornell University Ithaca, NY, 1989.

\bibitem[Howe and Cutkosky(1996)]{howe1996practical}
Robert~D Howe and Mark~R Cutkosky.
\newblock \href{https://journals.sagepub.com/doi/10.1177/027836499601500603}{Practical force-motion models for sliding manipulation}.
\newblock \emph{The International Journal of Robotics Research}, 15\penalty0 (6):\penalty0 557--572, 1996.

\bibitem[Xydas and Kao(1999)]{xydas1999modeling}
Nicholas Xydas and Imin Kao.
\newblock \href{https://journals.sagepub.com/doi/abs/10.1177/02783649922066673}{Modeling of contact mechanics and friction limit surfaces for soft fingers in robotics, with experimental results}.
\newblock \emph{The International Journal of Robotics Research}, 18\penalty0 (9):\penalty0 941--950, 1999.

\bibitem[Hogan and Rodriguez(2020{\natexlab{a}})]{hogan2020feedback}
Fran{\c{c}}ois~Robert Hogan and Alberto Rodriguez.
\newblock \href{https://link.springer.com/chapter/10.1007/978-3-030-43089-4_51}{Feedback control of the pusher-slider system: A story of hybrid and underactuated contact dynamics}.
\newblock In \emph{Algorithmic Foundations of Robotics XII}, pages 800--815. Springer, 2020{\natexlab{a}}.

\bibitem[Hogan and Rodriguez(2020{\natexlab{b}})]{hogan2020reactive}
Fran{\c{c}}ois~Robert Hogan and Alberto Rodriguez.
\newblock \href{https://journals.sagepub.com/doi/full/10.1177/0278364920913938}{Reactive planar non-prehensile manipulation with hybrid model predictive control}.
\newblock \emph{The International Journal of Robotics Research}, 39\penalty0 (7):\penalty0 755--773, 2020{\natexlab{b}}.

\bibitem[Kloss et~al.(2022)Kloss, Schaal, and Bohg]{kloss2022combining}
Alina Kloss, Stefan Schaal, and Jeannette Bohg.
\newblock \href{https://journals.sagepub.com/doi/full/10.1177/0278364920954896}{Combining learned and analytical models for predicting action effects from sensory data}.
\newblock \emph{The International Journal of Robotics Research}, 41\penalty0 (8):\penalty0 778--797, 2022.

\bibitem[Shi et~al.(2017)Shi, Woodruff, Umbanhowar, and Lynch]{shi2017dynamic}
Jian Shi, J~Zachary Woodruff, Paul~B Umbanhowar, and Kevin~M Lynch.
\newblock \href{https://ieeexplore.ieee.org/document/7913727}{Dynamic in-hand sliding manipulation}.
\newblock \emph{IEEE Transactions on Robotics}, 33\penalty0 (4):\penalty0 778--795, 2017.

\bibitem[Zhou et~al.(2016)Zhou, Paolini, Bagnell, and Mason]{zhou2016convex}
Jiaji Zhou, Robert Paolini, J~Andrew Bagnell, and Matthew~T Mason.
\newblock \href{https://ieeexplore.ieee.org/abstract/document/7487155}{A convex polynomial force-motion model for planar sliding: Identification and application}.
\newblock In \emph{2016 IEEE International Conference on Robotics and Automation (ICRA)}, pages 372--377. IEEE, 2016.

\bibitem[Gutowski and Leus(2012)]{gutowski2012effect}
Pawe{\l} Gutowski and Mariusz Leus.
\newblock The effect of longitudinal tangential vibrations on friction and driving forces in sliding motion.
\newblock \emph{Tribology International}, 55:\penalty0 108--118, 2012.

\bibitem[Hao et~al.(2018)Hao, Ping, Yang, and Tianshou]{hao2018effect}
Wu~Hao, Chen Ping, Liu Yang, and Ma~Tianshou.
\newblock Effect of axial vibration on sliding frictional force between shale and 45 steel.
\newblock \emph{Shock and Vibration}, 2018\penalty0 (1):\penalty0 4179312, 2018.

\end{thebibliography}

\newpage
% \appendix

% \section{Proof of motion}
% % \todo{prove the motion can be divided into translation/pure rotation}

\end{document}